\documentclass[11pt, a4paper, logo, twocolumn, internal, copyright, nonumbering]{googledeepmind}

\usepackage[authoryear, sort&compress, round]{natbib}
\usepackage{tcolorbox}
\tcbuselibrary{listings,breakable}
\usepackage{listings}

\usepackage{hyperref}
\usepackage{url}

\usepackage{algorithm}
\usepackage{algpseudocode}

\usepackage{booktabs}

\usepackage[english]{babel}
\usepackage{amsthm}
\newtheorem{theorem}{Theorem}[section]
\newtheorem{defn}[theorem]{Definition}

\usepackage{float}
\floatstyle{plaintop}
\restylefloat{table}
\usepackage[tableposition=top]{caption}

\usepackage{colortbl}

\usepackage{multirow}

\usepackage{wrapfig}

\usepackage{tablefootnote}

\usepackage{verbatim}

\usepackage{pifont}%
\newcommand{\cmark}{\ding{51}}%

\newcommand*\rot{\rotatebox{90}}

\usepackage{listings}
\usepackage{xcolor}
\definecolor{codegreen}{rgb}{0,0.6,0}
\definecolor{codegray}{rgb}{0.5,0.5,0.5}
\definecolor{codepurple}{rgb}{0.58,0,0.82}
\definecolor{backcolour}{rgb}{0.95,0.95,0.92}
\lstdefinestyle{mystyle}{
    backgroundcolor=\color{backcolour},   
    commentstyle=\color{codegreen},
    keywordstyle=\color{magenta},
    numberstyle=\tiny\color{codegray},
    stringstyle=\color{codepurple},
    basicstyle=\ttfamily\footnotesize,
    breakatwhitespace=false,         
    breaklines=true,                 
    captionpos=b,                    
    keepspaces=true,                 
    numbers=left,                    
    numbersep=5pt,                  
    showspaces=false,                
    showstringspaces=false,
    showtabs=false,                  
    tabsize=2
}
\definecolor{eclipseStrings}{RGB}{42,0.0,255}
\definecolor{eclipseKeywords}{RGB}{127,0,85}
\colorlet{numb}{magenta!60!black}
\lstdefinelanguage{json}{
    backgroundcolor=\color{backcolour},   
    commentstyle=\color{codegreen},
    keywordstyle=\color{magenta},
    numberstyle=\tiny\color{codegray},
    stringstyle=\color{codepurple},
    basicstyle=\ttfamily\footnotesize,
    numbers=left,
    numberstyle=\scriptsize,
    stepnumber=1,
    numbersep=8pt,
    showstringspaces=false,
    breaklines=true,
    string=[s]{"}{"},
    comment=[l]{:\ "},
    morecomment=[l]{:"},
    literate=
        *{0}{{{\color{numb}0}}}{1}
         {1}{{{\color{numb}1}}}{1}
         {2}{{{\color{numb}2}}}{1}
         {3}{{{\color{numb}3}}}{1}
         {4}{{{\color{numb}4}}}{1}
         {5}{{{\color{numb}5}}}{1}
         {6}{{{\color{numb}6}}}{1}
         {7}{{{\color{numb}7}}}{1}
         {8}{{{\color{numb}8}}}{1}
         {9}{{{\color{numb}9}}}{1}
}

\title{PRISM: Efficient Long-Range Reasoning With Short-Context LLMs}

\correspondingauthor{dulhan@robots.ox.ac.uk, bgunel@google.com}


\renewcommand{\today}{2025-08-24}

\renewcommand{\internalonly}{Preprint.}
\renewcommand{\copyrightext}{*Work done while author was at Google DeepMind.}

\author[*,1]{Dulhan Jayalath}
\author[2]{James B. Wendt}
\author[2]{Nicholas Monath}
\author[2]{Sandeep Tata}
\author[2]{Beliz Gunel}

\affil[1]{University of Oxford}
\affil[2]{Google DeepMind}

\begin{abstract}
Long-range tasks demand reasoning over long inputs. However, existing solutions are limited, e.g., long-context models require large compute budgets, parameter-efficient fine-tuning (PEFT) needs training data, and retrieval-augmented generation (RAG) entails complex task-specific designs. Though in-context approaches overcome many of these issues, methods with short-context LLMs are inefficient, trading context for processing more tokens. We introduce \textbf{PRISM}, a highly token-efficient in-context method based on structured schemas that outperforms baselines on diverse tasks with \textbf{4x shorter contexts}. This approach produces concise outputs and efficiently leverages key-value (KV) caches to \textbf{reduce costs by up to 54\%}. PRISM scales down to tiny contexts without increasing costs or sacrificing quality, and generalizes to new tasks with minimal effort by generating schemas from task descriptions.
\end{abstract}

\begin{document}

\maketitle

\section{Introduction}

Long information contexts pose significant challenges for language tasks. The prototypical example is long document summarization, where a lengthy piece of text must be summarized into a short-form summary. For these and other long natural language tasks, large language models (LLMs) are state-of-the-art. In summarization, an LLM is typically prompted with summarization instructions alongside the text and generates a summary of the content. However, this requires sufficient context to accommodate the entire document. Many practitioners and researchers rely on models with short contexts because they are limited by the inference cost of long-context models, open source or on-premises requirements, local compute constraints, or other barriers. There are a range of alternatives to long-context language models, which include PEFT and RAG. However, these solutions either require training data or necessitate complex task-specific design choices. Short context LLMs with in-context methods promise a training data-free and task-agnostic alternative to solving long-range tasks. They achieve this by repeatedly applying short context models to chunks of long text, requiring processing a very large number of input and output tokens. In turn, this leads to high API usage or compute costs. In response, we design \textbf{PRISM}, \textit{Processing Incrementally with Structured Memory}: a highly token-efficient in-context approach that is task-agnostic, requires no training data, uses a small compute budget, and does not need access to model weights. No existing method satisfies all these constraints.

\begin{figure}
\begin{center}
    \includegraphics[width=1.0\linewidth]{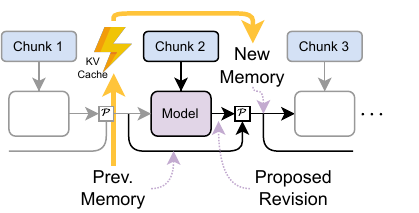}
\caption{\textbf{PRISM} efficiently processes a stream of chunked data, leveraging a concise and cache-optimized structured memory to propose revisions at each step.}
\vspace{-25pt}
\label{fig:incremental}
\end{center}
\end{figure}

PRISM employs incremental processing, treating the input as a sequential stream of chunks, processed in the order of their appearance alongside a structured in-context memory of prior chunks. While incremental methods are not new, e.g., \citet{bookscore2024, cok2024, Qian2024AreLA}, existing approaches are task-specific and are not economic in terms of tokens processed. PRISM specifically addresses these limitations through an optimized structured memory. Rather than seeing a natural language memory, the LLM leverages a structured representation of prior information and outputs a proposed revision to the memory based on the current chunk (Figure \ref{fig:incremental}). The memory is specified by a user-defined typed hierarchical schema, supporting any kind of long-range task. PRISM uses the structured memory to track salient prior information more succinctly than with natural language. Instead of the output of the LLM overwriting the memory, it proposes a structured revision which is used to programmatically revise the memory. This design yields concise outputs and reduces the reasoning burden on the LLM. Crucially, we design the memory to efficiently leverage prefix key-value caching \citep{pagedattention2023, scalinginference2023} by intelligently reusing activations computed from unchanged memory segments in prior steps. Taken together, our approach yields both higher quality results and greater token efficiency. 

\noindent Our main contributions are: (1) \textbf{PRISM}: an approach for \textbf{solving long-range tasks} with better quality (Section \ref{sec:quality}), efficiency (Section \ref{sec:efficient}), and fewer constraints (Table \ref{tab:comparison}) than alternatives; (2) an empirical analysis demonstrating PRISM's \textbf{token efficiency} and \textbf{scalability} to shorter chunks (Section \ref{sec:efficient}); and (3) evidence PRISM generalizes to new tasks with \textbf{generated schemas} (Section \ref{sec:llm-schema}).

\begin{table}
    \centering
    \begin{tabular}{lcccc}
        \textbf{Method} & \rot{No training} & \rot{No weights} & \rot{Low compute} & \rot{Task-agnostic} \\
        \midrule
        Long-context models & \cmark & \cmark & & \cmark \\
        RAG & \cmark & \cmark & \cmark & \\
        PEFT & & & \cmark & \\
        In-context alternatives & \cmark & \cmark & & \textbf{?} \\
        \textbf{PRISM (Ours)} & \cellcolor{green!25}\cmark & \cellcolor{green!25}\cmark & \cellcolor{green!25}\cmark & \cellcolor{green!25}\cmark \\
        \bottomrule
    \end{tabular}
    \caption{\textbf{Comparison of approaches for long-range tasks.} While existing methods each have limitations, PRISM satisfies all constraints: it requires no training data, needs no access to model weights, operates within a low compute budget, and remains task-agnostic, making it suitable for a wide range of applications.}
    \label{tab:comparison}
\end{table}

\section{Related Work}

Several approaches tackle long-range reasoning with limited contexts through various memory-based and memory-less mechanisms. For document processing, methods include hierarchical summarization \citep{bookscore2024}, natural language knowledge representations \citep{sumie2024}, sequential scanning and selection \citep{Qian2024AreLA}, and JSON-encoded memories \citep{cok2024}. \citet{retrievalreasoning2024} specifically target retrieval-based question-answering and \citet{memgpt2024} perform stateful reasoning in LLMs using function calls to read and write data to storage. While some of these methods address specific domains, they lack PRISM's general applicability across task types and, crucially, all neglect the token efficiency that PRISM achieves.

In contrast to external memories, another research direction embeds memories directly into model architectures. Several works transform LLMs into recurrent models through memory embeddings \citep{hmt2024} or latent-space memories \citep{infiniattention2024}. \citet{wang2023} design a new model architecture with a retrieval side-network to cache and update a memory and \citet{ivgi2023} propose using a language model encoder to encode overlapping chunks and fuse information with a pre-trained decoder. Unlike these methods requiring architectural modifications or weight access, PRISM maintains a structured external memory that works with any black-box LLM.

By using a structured schema to organize information and optimizing it for KV caches, PRISM achieves both task-agnosticism and token efficiency without requiring model modifications or training—addressing core limitations of prior work.

\section{Method}

We seek to solve long-range tasks token-efficiently without long-context models. By using an incremental processing strategy with a structured memory, we resolve many of the constraints of other methods (Table \ref{tab:comparison}). In this section, we define the incremental processing strategy, provide a way to structure the memory using a typed hierarchical schema, and show how to efficiently process these tokens across multiple LLM calls.

\subsection{Incremental Processing Formulation}
In the incremental view, instead of seeing the entire context at once, the LLM sees contiguous segments (which we refer to as \textit{chunks}) in sequence. To avoid forgetting previous information, the LLM also sees a memory, encoding information about prior chunks relevant to the task. This memory is constructed from the output of the LLM in the previous step. In the current call, the LLM uses its output to revise the memory based on the information in the current chunk. The use of LLM outputs as a memory in this way is characteristic of solving incremental tasks using LLMs.

Formally, data arrives in increments, forming an ordered sequence of chunks $(d_1, d_2, \dots, d_n)$. An LLM is prompted over multiple incremental steps $i \in \{1, \dots, n\}$, with task instructions $\mathcal{T}$, the next chunk $d_i$, and the output of the model from the previous step $o_{i-1}$. Accordingly, the prompt is a tuple $(\mathcal{T}, d_i, o_{i-1})$. The output of the previous step acts as a natural language memory that assists in solving the task. This implies a definition for an in-context memory:
\begin{defn} 
An in-context memory is the tokens input to the model in an incremental step that encode the prior information seen by the model.
\label{ref:def1}
\end{defn}
\noindent In this formulation, the LLM revises the memory by overwriting it through the tokens it decodes in the next incremental step, forming the next state of the memory. The output of the final step $o_n$ is taken as the answer or otherwise post-processed.

\subsection{Using Structured Memories}
\label{sec:main-method}

Natural language (or \textit{unstructured}) memories do not necessarily encode the most salient information for a specific task because the output format is unconstrained. This often impairs task performance. We improve the typical incremental processing formulation by introducing a structured memory and structured output to increase information relevance and reduce the LLM's cognitive burden.

To introduce a structured constraint in PRISM, we replace the natural language memory with a structured memory. Specifically, we prompt the language model at step $i$ with a modified tuple $(\mathcal{T}, \mathcal{S}, m_i, d_i)$ where we replace the natural language memory $o_{i-1}$ with a structured memory $m_i$ specified by a typed hierarchical schema.

\begin{defn}
A typed hierarchy is a structure of primitive types and simple data structures (e.g., integers, floating points, strings, and lists) in a hierarchy laid out by nested key-value maps.
\end{defn}
\begin{defn}
A structured memory $m$ has a schema $\mathcal{S}$ specified with a typed hierarchy.
\end{defn}

\noindent For example, a simple schema for narrative summarization could be (with Python-like typing) \texttt{str:} \texttt{list<str>} i.e., a key-value mapping of character names to strings describing events involving that character. After seeing a new story chunk, we can revise information about a character by adding to the entries for that character. We choose to use typed hierarchies because they are easily addressable (using a path specified by keys and indices) and updatable. We specify a new schema for each task as this structure determines the information that will be encoded in the memory.

To revise the memory, instead of generating a structured memory to overwrite the prior memory, the output of the model is a proposed memory revision $r_i$, which provides a path to programmatically revise $m_i$ with a new value. Proposing revisions rather than overwriting the entire memory saves tokens and improves efficiency.

\begin{defn}
A structured memory revision $r$ is a tuple $(p, o, v)$ where $p$ specifies an addressable path in the memory, $o$ is a binary operation that is either \texttt{add} or \texttt{update} and $v$ is the value to revise the memory with.
\label{defn:update}
\end{defn}

\begin{figure}[h]
    \centering
    \includegraphics[width=1.0\linewidth]{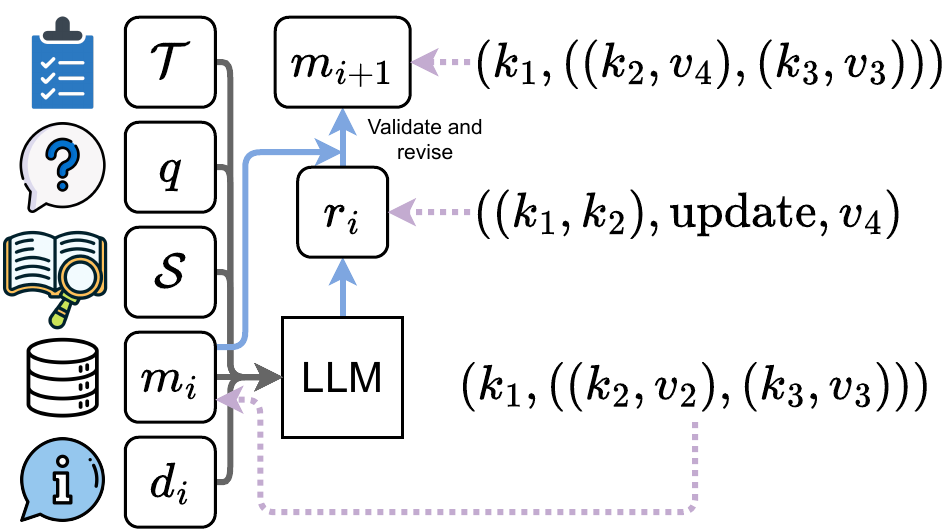}
    \caption{\textbf{PRISM with typed examples.} The model receives as input the tuple $(\mathcal{T}, q, \mathcal{S}, m_i, d_i)$ describing the task, query, schema, the current state of the memory, and the current chunk. The model outputs a proposed revision $r_i$ to programmatically revise the memory state to $m_{i+1}$. The purple arrows annotate example memory and revision states. Here, $v_2$ in $m_i$ is replaced with $v_4$ in $m_{i+1}$ using the path, operation, and value in the revision. If the operation were \texttt{add} instead, then the path would be created and $v_4$ added to the memory. To instead update $k_3$ with $v_4$, the addressable path would be $(k_1, k_3)$, leading to output revision $((k_1, k_3), \mathrm{update}, v4)$.
    }
    \label{fig:overview}
\end{figure}

\begin{figure}[h]
    \centering
    \includegraphics[trim={2.8cm 0 0 0},clip,width=1.0\linewidth]{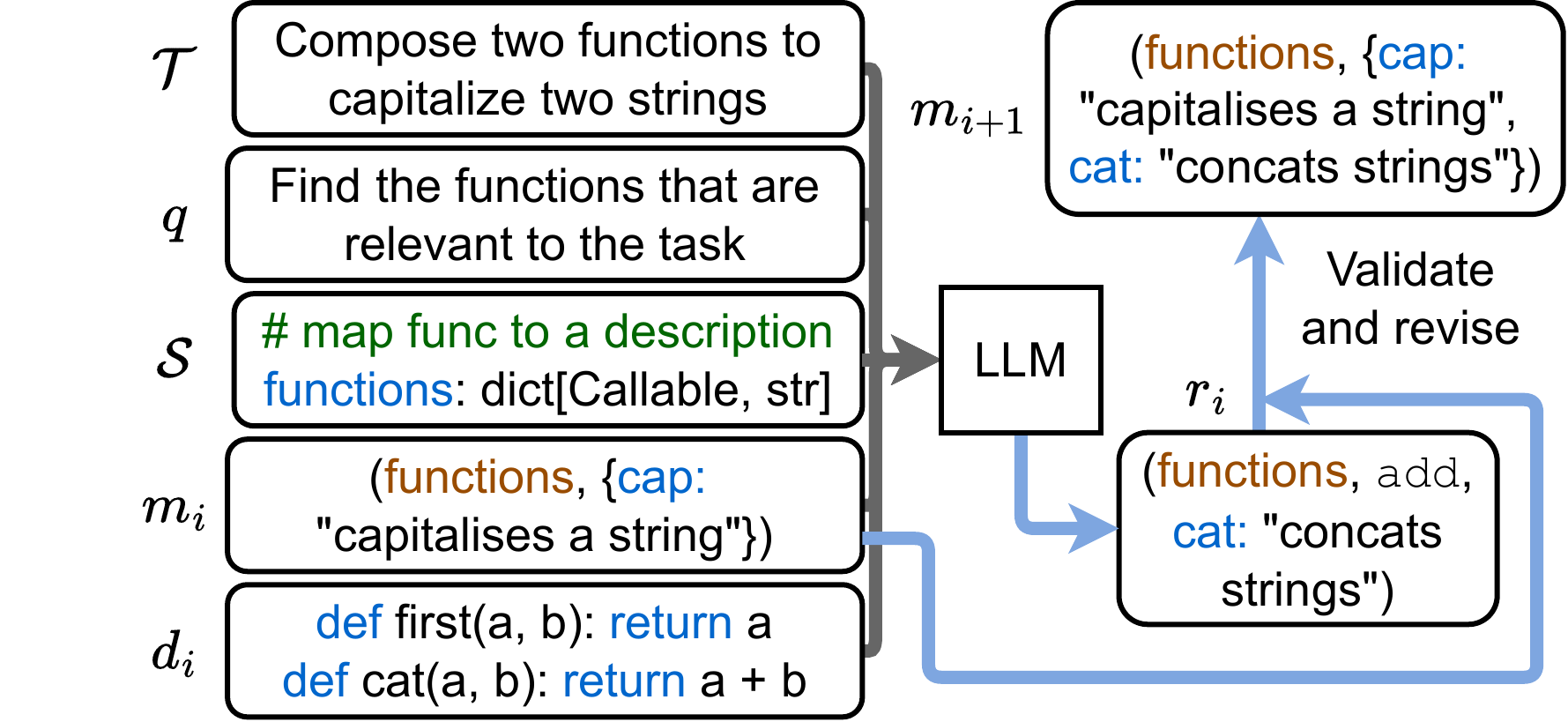}
    \caption{\textbf{PRISM code composition example.} The LLM proposes adding the \texttt{cat} function from the chunk $d_i$ (and a description) to the existing memory because it best fits the query. The memory now has \texttt{cap} and \texttt{cat}.}
    \label{fig:prism-example}
\end{figure}

\noindent If $o$ is \texttt{add}, $p$ specifies a new path to which $v$ is added; if \texttt{update}, $p$ specifies an existing path in the memory whose current value should be replaced with $v$. After validating the proposed revision by programmatically ensuring it conforms to the expected structure, the memory $m_i$ is revised with $r_i$ to the next memory state $m_{i+1}$. Figure \ref{fig:overview} provides an overview of our approach and Figure \ref{fig:prism-example} gives a concrete example. In practice, $r_i$ may consist of more than one proposed revision.

After processing all chunks, the LLM uses the final state of the memory (alongside the query and a specification of the memory structure) to give a final answer. Algorithm \ref{alg:method} shows all steps.

\begin{algorithm}
\caption{PRISM}
\begin{algorithmic}[1]
\Require $\mathcal{T}$, $q$, $\mathcal{S}$, $(d_1, d_2, \dots, d_n)$ \Comment{Task instruction, query, memory schema, and chunks of information}
\State $m_1$ $\gets$ $\{\}$
\For{$i = 1$ \textbf{to} $n$} 
    \State $r_i$ $\gets$ LLM($\mathcal{T}$, $q$, $\mathcal{S}$, $m_i$, $d_i$)
    \State $m_{i+1}$ $\gets$ ReviseMemory($m_i$, $r_i$) \Comment{Add to or update the memory with the proposed value}
\EndFor
\State answer $\gets$ LLM($\mathcal{T}_{\mathrm{final}}$, $q$, $\mathcal{S}$, $m_{n+1}$) \Comment{Generate the answer using the final memory}
\State \Return answer
\end{algorithmic}
\label{alg:method}
\end{algorithm}

\noindent Our approach brings several quality benefits. First, a structured memory constrains the output to the query domain. This gives the model focus by forcing it to generate only the information we have deemed relevant for the query (via the schema $\mathcal{S}$) to revise the memory. Having a structured memory also assists the LLM in understanding and updating relevant information for the task. By using a structured memory, we provide flexibility in deciding how to construct the memory structure for a particular type of task or to even automate the generation of the schema. Furthermore, we output a \textit{revision} (i.e., the difference between the current and next memory state) rather than the memory itself, reducing the number of tokens to decode. Beyond the quality benefits, our structured approach enables significant token efficiency gains through key-value cache utilization, PRISM's core contribution, which we explore next.

\subsection{Token-Efficient Memory Encoding}

Encoding memories increases token count and processing time. This can become a significant bottleneck when there are many chunks of information in an incremental task or if the size of the memory dominates the rest of the prompt. 

One way to improve encoding efficiency is to utilize prefix KV caching \citep{radix2023} to store and reuse previously computed token activations. With this method, if there is a prefix of the prompt that matches a prior encoded prompt, the model can reuse the KV activations previously computed for this prefix. Thus, maximizing the length of this prefix is essential for cache efficiency.  For simplicity, our experiments implement prefix KV caching such that the KV activations are reused for only the longest prefix matching the \textit{last} encoded prompt. Most prefix caching implementations will store activations from further in the past and may lead to even higher cache efficiency as a result of being able to retrieve matching prefixes from multiple past prompts.

To leverage the cache utilization improvements we introduce next, we first ensure that our prompt is KV cache-friendly. The prompt is the tuple $(\mathcal{T}, \mathcal{S}, m_i, d_i)$. Since only $m$ and $d$ will change between incremental steps, there is no need to re-encode the tokens for the prefix ($\mathcal{T}$, $\mathcal{S}$). We arrange the prompt so that the memory $m$ appears \textit{before} the chunk $d$ rather than after because while the tokens in the chunk will likely be different, parts of the memory may not change across steps. As our method produces memory revisions, which do not necessarily always overwrite the entire memory, key-value activations can be reused when encoding memory $m_i$ up until the point of the first change to the memory from the previous prompt $m_{i-1}$. Reusing a substantial number of token activations would be unlikely in the usual problem formulation with natural language memories.

\begin{figure}
    \centering
    \includegraphics[width=1.0\linewidth]{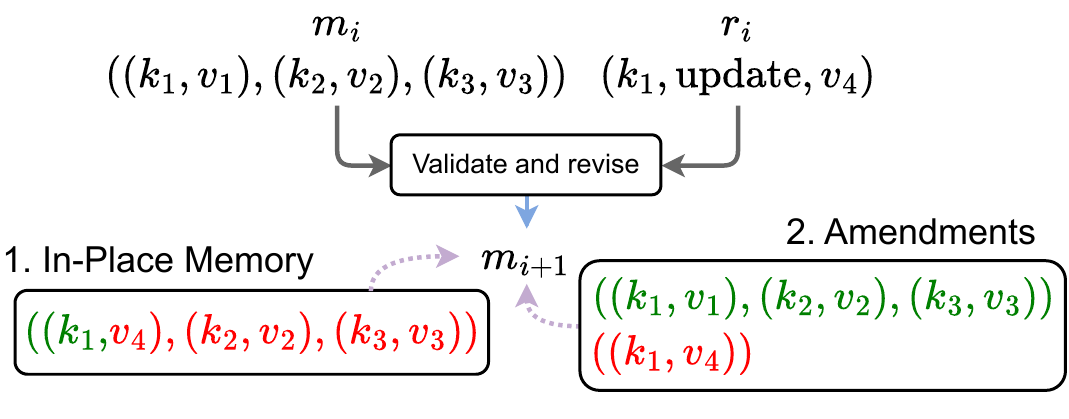}
    \caption{\textbf{PRISM's amendments improve KV cache utilization.} Using a single-level key-value map as the memory $m_i$, we show an update proposed to the value at $k_1$. After applying it, we get memory state $m_{i+1}$ which can be represented in one of two ways: an \textit{in-place memory} where the value is updated directly or as \textit{amendments} where the change is amended to the end of the memory state as a new memory structure. Green shows the longest matching prefix compared to the previous memory $m_i$ and red shows the information that must be (re-)encoded. Using amendments reduces the number of tokens that need to be encoded at the cost of increasing the size of the memory.
    }
    \label{fig:amendments}
\end{figure}

We now introduce \textit{amendments}, a novel approach to maximize cache utilization. If, instead of updating the path $p$ in the memory with the new value $v$, we add a new memory, which we call an \textit{amendment}, containing the new value and its path directly after the existing one, then the KV activations for everything up to the newest change can always be reused. This requires the LLM to reason more about the memory by understanding that subsequent amendments with existing paths overwrite prior paths. Adding \textit{amendments} is an alternative to maintaining an \textit{in-place} memory (Figure \ref{fig:amendments}) and creates a configurable efficiency trade-off. Amendments reduce encoding costs but may increase memory tokens if update operations, rather than additions, dominate. A choice between these options should be made based on the expected operation patterns of a specific task, optimizing for the best performance in each scenario.

\subsection{Generating Memory Schemas}

To minimize implementation effort and expand domain coverage, the memory schema can be automatically generated by prompting an LLM. We hand-craft three schemas from a variety of domains, using these as few-shot examples, and prompt the LLM (Appendix \ref{app:schemagen}) to generate a schema for a \textit{different} domain given a simple description of the query domain and an example query.

For example, if the task is code retrieval, the prompt should describe the query domain, the task of retrieving a function given a code repository, and provide an example query which describes the procedure of a function as well as its inputs and outputs. The output of the LLM is then a schema that defines the structure of a memory that encodes information relevant to this task from the chunks seen by the LLM. This could be something like a map from the names of functions seen to a brief description of what the function does.

Other than automatically generating schemas reducing human effort, we hypothesize that an LLM can produce a more relevant schema for a task than what a non-expert may construct. Thus, schema generation makes PRISM accessible beyond domain specialists.

\section{Experiments}

\textbf{Datasets}\quad We use three state-of-the-art long-range datasets spanning the spectrum of reasoning tasks. \textit{BooookScore} \citep{bookscore2024} is both a long-context book summarization dataset and benchmark metric. It contains very large books (each over 100k tokens) curated to ensure they did not exist in the data of public LLMs at the time of publication. \citet{bookscore2024} also introduce a reference-free summarization metric with the same name which we use to measure the coherency of summaries. This is an LLM-assisted measure that computes the 
\% of sentences in the summary that do \textit{not} contain any of a number of error types. The second dataset is a long-range code understanding and retrieval benchmark called \textit{RepoQA} \citep{repoqa2024}. Inputs are large code repositories totalling above 100k tokens. The task is to retrieve a function, described in natural language without being named, from the repository. A memory is useful to reason about this task because function descriptions describe behavior through relationships with other functions. We measure accuracy, marking an output as correct if it names the described function exactly. Our final task, which we refer to as \textit{LOFT-Spider} \citep{loft2024}, requires answering a set of questions directly (rather than via SQL commands) from a large SQL database. Response accuracy on this task is measured using exact match. These datasets evaluate opposing boundaries in LLM reasoning. BooookScore is an unstructured natural language reasoning task, while RepoQA and LOFT-Spider are well-structured retrieval and reasoning tasks.

\noindent \textbf{Models}\quad To establish a quality ceiling, we compare our baselines to a state-of-the-art long context model, Gemini 1.5 Pro \citep{gemini2024}, with a context of 1M tokens. This is large enough to fit the longest samples from each of the datasets we study \textit{within} context. For all other baselines, we use the same model with 32k context, isolating the impact of context length while keeping model capability constant. We use top $k$ sampling ($k=40$) with temperature $0.8$. 

\noindent \textbf{Baselines}\quad We use \textit{incremental} merging and \textit{hierarchical} merging as our short-context baselines for BooookScore. These were proposed by \citet{bookscore2024} alongside the dataset. Incremental merging follows the characteristic incremental task formulation of revising a running summary in natural language as new chunks are seen; hierarchical merging summarizes all chunks, then summarizes consecutive pairs of summaries hierarchically in layers until a single summary remains at the last layer. As RepoQA lacks short-context baselines, we adapted the incremental merging approach from \citet{bookscore2024}, modifying prompts to suit the retrieval task. We also construct a similar baseline for LOFT-Spider. A hierarchical baseline is not naturally amenable to these latter tasks as it is unclear how to merge summaries of independent functions or tables nor why it would be beneficial.

\noindent \textbf{Ablations}\quad To isolate components of our approach, we evaluate several variations of our method. We compare in-place memories to amendments (Figure \ref{fig:amendments}) to see the effect of caching improvements. We also evaluate when the proposed revision (Definition \ref{defn:update}) supports both \texttt{add} and \texttt{update} as well as when it supports only the \texttt{add} operation since using only the \texttt{add} operation can reduce the number of tokens decoded.

\noindent \textbf{Setup}\quad We encode our typed hierarchy in \textit{JavaScript Object Notation (JSON)} and specify the schema for the memory using Python 3 dataclasses. Appendix \ref{sec:json} provides some examples of this implementation. Unless specified otherwise, we use the schemas defined in Appendix \ref{sec:handschema}. We evaluate on 50 examples per dataset due to compute restrictions, and report the mean of the dataset-specific metric over all samples. We quote uncertainty as the standard error of the mean over five solutions generated through our method.

\subsection{PRISM Outperforms Alternatives While Using Shorter Contexts}
\label{sec:quality}

\begin{table*}[t]
    \centering
    \begin{tabular}{l|rr|rr|rr}
    \toprule
    & \multicolumn{2}{c|}{\textbf{BooookScore}} & \multicolumn{2}{c|}{\textbf{RepoQA}} & \multicolumn{2}{c}{\textbf{LOFT-Spider}} \\
    \cmidrule(lr){2-3} \cmidrule(lr){4-5} \cmidrule(lr){6-7}
    \textbf{Method} & Score & Ch. Tokens & Acc. & Ch. Tokens & Acc. & Ch. Tokens \\
    \midrule
    Long context & $.67 {\pm} .004$ & 100k & $.92$ & 121k & $.44 {\pm} .01$ & 30k \\
    \midrule
    Baselines$^\dagger$ & & & & & & \\
    ~~Incremental & $.63 {\pm} .010$ & \multirow{2}{*}{--} & $.24 {\pm} .03$ & \multirow{2}{*}{--} & $.12 {\pm} .02$ & \multirow{2}{*}{--} \\
    ~~Hierarchical & $.51 {\pm} .006$ & & \multicolumn{2}{c|}{n/a} & \multicolumn{2}{c}{n/a} \\
    \midrule
    \textbf{PRISM} & & & & & & \\
    ~~In-place & $.63 {\pm} .008$ & \multirow{2}{*}{\textbf{2k}} & $.42 {\pm} .02$ & \multirow{2}{*}{\textbf{8k}} & $.22 {\pm} .03$ & \multirow{2}{*}{\textbf{8k}} \\
    ~~~~+ w/o updates & $.63 {\pm} .006$ & & $.50 {\pm} .03$ & & $\mathbf{.26} {\pm} .02$ & \\
    ~~Amendments & $\mathbf{.65} {\pm} .004$ & \multirow{2}{*}{\textbf{2k}} & $.48 {\pm} .03$ & \multirow{2}{*}{\textbf{8k}} & $.14 {\pm} .01$ & \multirow{2}{*}{\textbf{8k}} \\
    ~~~~+ w/o updates & $.63 {\pm} .005$ & & $\mathbf{.53} {\pm} .02$ & & $.22 {\pm} .02$ & \\
    \bottomrule
    \multicolumn{7}{l}{\small $^\dagger$Baselines adapted from \citet{bookscore2024}'s methods.} \\
    \end{tabular}
    \caption{\textbf{PRISM closes the gap between baselines and long-context models using 4-50x smaller context windows.} Using just 2-8k token chunks (vs. 30-121k for long context), PRISM significantly outperforms baselines across all tasks. On BooookScore, PRISM's amendments achieve 97\% of long-context performance ($p{<}.02$) with 50x smaller context. On RepoQA, PRISM reaches 58\% of ceiling accuracy while using 15x less context.}
    \label{tab:quality}
\end{table*}

In Table \ref{tab:quality}, our method beats both existing baselines (incremental and hierarchical merging) on all datasets to a statistically significant degree (at worst $p=0.02$). This suggests that our structured approach and memory provide meaningful improvements in reasoning performance over alternatives. This could be a result of constraining the LLM to produce outputs that are directly relevant to the task using the structured memory. We also note that our approach generally benefits or performs on par with in-place memories when using amendments instead. This is a promising signal that cache-optimized memories can be just as effective at producing strong final answers.

In all datasets, PRISM begins to approach the long context ceiling and in BooookScore, it almost matches it. RepoQA and LOFT-Spider are more difficult tasks that necessitate reasoning and aggregating over multiple code files and tables. It is not trivial to define a schema that optimally supports the reasoning involved. We also believe that critical information is clustered in these tasks and failing to add relevant information to the memory during the processing of an important chunk is likely to be substantially more costly than in summarization.

While PRISM achieves strong performance across all tasks with significantly smaller context windows, what is the computational cost? Traditional memory approaches often trade increased token usage for improved reasoning capabilities. In the following section, we demonstrate that PRISM not only improves performance but does so with substantial efficiency gains.

\begin{table*}[t]
    \centering
    \begin{tabular}{l@{\hspace{4pt}}l|rrrrr}
    \toprule
    \multicolumn{2}{l|}{\multirow{2}{*}{\textbf{Method}}} & Cache & \multicolumn{2}{c}{Tokens ($\times10^3$)} & Output & Cost \\
    \cmidrule(lr){4-5}
    & & Hit (\%) & Total & Net & ($\times10^3$) & Index \\
    \midrule
    \multicolumn{7}{c}{\textit{BooookScore}} \\
    \midrule
    \multicolumn{2}{l|}{Long context} & -- & 100 & 100 & 1 & \cellcolor{gray!10}-- \\
    \addlinespace[4pt]
    \multicolumn{2}{l|}{Baselines} & & & & & \\
    \multicolumn{2}{l|}{~~Incremental$^\dagger$} & 0 & 249 & 248 & 141 & 0.67 \\
    \multicolumn{2}{l|}{~~Hierarchical$^\dagger$} & 1 & 227 & 225 & 70 & 0.43 \\
    \addlinespace[4pt]
    \textbf{PRISM} & & & & & & \\
    ~~In-place & & 49 & 495 & 250 & 131 & 0.64 \\
    ~~~~+ w/o updates & & 34 & 619 & 409 & \textbf{14} & 0.45 \\
    ~~Amendments & & \textbf{69} & 559 & \textbf{171} & 47 & \textbf{0.31} \\
    ~~~~+ w/o updates & & 37 & 676 & 424 & 15 & 0.47 \\
    \midrule
    \multicolumn{7}{c}{\textit{RepoQA}} \\
    \midrule
    \multicolumn{2}{l|}{Long context} & -- & 121 & 121 & 0 & \cellcolor{gray!10}-- \\
    \multicolumn{2}{l|}{Incremental} & 1 & 180 & 178 & 28 & 0.26 \\
    \addlinespace[4pt]
    \textbf{PRISM} & & & & & & \\
    ~~In-place & & 71 & 491 & 142 & 9 & 0.17 \\
    ~~~~+ w/o updates & & 68 & 437 & 139 & \textbf{6} & 0.16 \\
    ~~Amendments & & \textbf{75} & 581 & 144 & 11 & 0.18 \\
    ~~~~+ w/o updates & & 68 & 435 & \textbf{138} & \textbf{6} & \textbf{0.15} \\
    \bottomrule
    \multicolumn{7}{l}{\small $^\dagger$Methods from \citet{bookscore2024}.} \\
    \multicolumn{7}{l}{\small Cost Index = (Net Tokens + 3 × Output) ÷ $10^6$, reflecting typical API pricing ratios.}
    \end{tabular}
    \caption{\textbf{PRISM with amendments achieves 69\% cache reuse and 54\% cost reduction.} 
    On BooookScore, PRISM's amendment strategy maximizes cache hits (69\% vs. 0-1\% for incremental and hierarchical baselines) 
    while minimizing net tokens (171k vs. 248k) and cost (0.31 vs. 0.67). Without updates further reduces output tokens by 70\% but increases net encoding. Similar patterns hold for RepoQA, where amendments achieve 75\% cache reuse. We do not provide a cost index for long context as the cost is different and would not be comparable. Results are similar for LOFT-Spider and shown in Appendix \ref{app:more} for brevity.}
    \label{tab:efficiency}
\end{table*}

\subsection{PRISM Is Scalable and Token-Efficient}
\label{sec:efficient}

In this section, we measure cache hit rate as the proportion of tokens whose key-value activations could be reused (i.e., the number of tokens in the longest matching prefix to the last input prompt divided by the total number of tokens encoded). We also compute a cost index to compare the relative compute cost of each method.

In Table \ref{tab:efficiency}, we see that variants of our method achieve the best results for all metrics across both datasets. Notably, using amendments with updates leads to the highest cache hit rates and generally cuts costs in half. In the case of BooookScore, it leads to the lowest estimated cost. Similarly, for RepoQA, using amendments (but this time without updates) leads to the lowest cost.

\begin{figure*}
    \centering
    \includegraphics[width=1.0\linewidth]{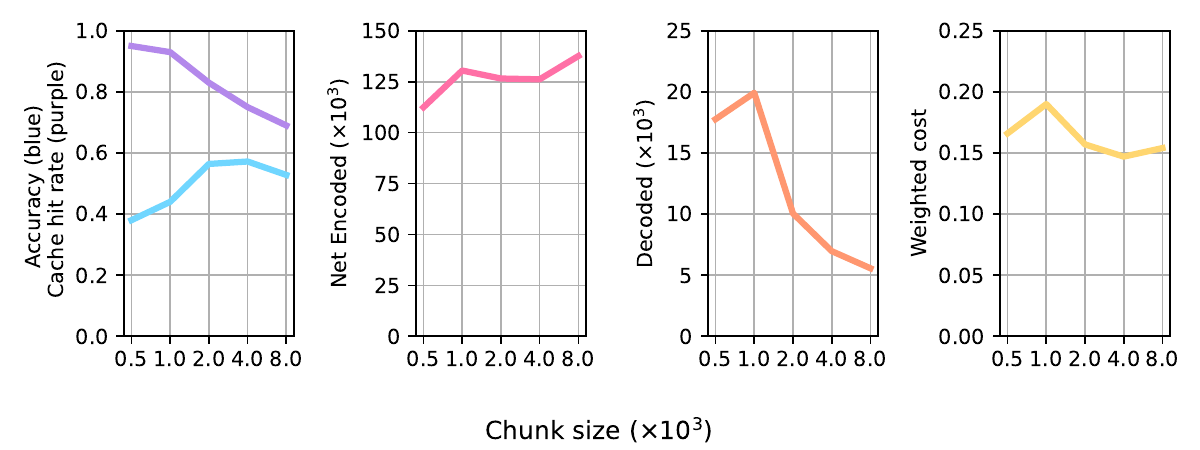}
    \caption{\textbf{PRISM costs do not increase with shorter chunk sizes due to effective KV caching.} This allows PRISM to scale down without sacrificing performance. Net tokens encoded are calculated after subtracting tokens reused. Weighted cost reflects typical API pricing (encoding + 3x decoding).}
    \label{fig:chunksize}
\end{figure*}

As compute constraints can significantly reduce viable context lengths, we also analyze how the characteristics of our method change as we reduce the context size. In Figure \ref{fig:chunksize}, we use different chunk sizes on the RepoQA dataset using our cache-efficient amended memory approach without updates. For larger chunk sizes, accuracy slightly increases. The net encoded tokens stays relatively constant since cache hit rate decreases. This is because a smaller proportion of the context is taken by the memory. Meanwhile, tokens decoded decreases as fewer incremental steps are required. However, tokens encoded dominates tokens decoded. Thanks to this property, remarkably, PRISM maintains consistent cost even as chunk sizes decrease by $4\times$. Thus, smaller chunks do not always lead to higher costs with our method.

\subsection{PRISM Works With Generated Schemas}
\label{sec:llm-schema}

In Table \ref{tab:schema}, we use LLM-generated schemas (Appendix \ref{sec:llmschema}) constructed by providing a brief description of the task and an example query (alongside some examples for other tasks) to the LLM. The output is a schema that we use to specify the memory. We compare this to the best result using our hand-crafted schemas from Table \ref{tab:quality}. The results reveal that our approach is competitive with hand-crafted expert schemas. Our method can be applied to tasks with little human input or domain expertise. For LOFT-Spider, we believe it is generally unclear what an optimal memory representation would be, and it is impressive that a strong representation can be constructed with an LLM.

\begin{table}[t]
    \centering
    \small
    \begin{tabular}{l|rrr}
    \toprule
    \textbf{Schema} & \textbf{B.Score} & \textbf{RepoQA} & \textbf{LOFT-Spider} \\
    \midrule
    Manual & $.65 \pm .004$ & $.53 \pm .02$ & $.26 \pm .02$ \\
    Gen. & $.61 \pm .010$ & $.53^\dagger \pm .02$ & $.15 \pm .03$ \\
    \bottomrule
    \end{tabular}
    \vspace{-2mm}
    \caption{\textbf{LLM-generated schemas match or approach experts.} Generated schemas achieve identical performance on 
    RepoQA, near-parity on BooookScore (93\% of expert), with some 
    limitations on Spider (58\% of expert). Both maintain similar 
    efficiency (Manual: 760 tokens, Generated: 790 tokens). 
    $^\dagger$Generated schema matched manual; alternative: $.24 \pm .01$.}
    \label{tab:schema}
\end{table}

\section{Conclusion \& Future Work}
PRISM demonstrates that structured in-context memories with programmatic revisions enable short-context models to match or approach long-context performance at substantially lower computational cost with high token-efficiency. We achieved better long-range task performance than baselines with unstructured memories for unstructured tasks, such as narrative summarization, as well as structured reasoning problems for code and databases. Our method was even competitive with a long-context model. We also demonstrated that PRISM is task-agnostic, requiring only specifying an appropriate schema for our memory. This too can be automated by generating the schema with an LLM. These LLM-assisted schemas achieved similar performance to expert schemas. Furthermore, with a slight modification to the memory representation, we improved key-value cache efficiency to reduce inference cost substantially below baselines without sacrificing task performance. Finally, we noticed that our method scales down without significantly increasing the inference cost and while remaining practical for long-range reasoning. Taken collectively, our method provides a solution for long-range reasoning without expensive long-context models and specialized methods.

We suggest several research directions for future work: (1) combine prior and future context rather than relying on incremental solutions; (2) applying our method to hierarchical memory approaches that capture both fine-grained details and high-level abstractions; (3) explore dynamically updating the schema based on incoming content; and (4) experiment with multi-stage PRISM processing, where the entire memory is revisited after all chunks are processed. This is computationally cheap due to our method's memory caching advantages.

\section{Limitations}
While we have shown that structured memories can be task-agnostic, improve quality, and improve token-efficiency, there remain some limitations to our work. First, we explore only three hand-crafted schemas across our tasks. There is likely to be a large space of effective and useful schemas for various types of tasks. Understanding how schemas should be designed for different tasks would help use structured memories more effectively.

Second, the analysis of chunk size and token efficiency is an interesting preliminary study that demonstrates the cost-efficiency of our approach even in increasingly context-constrained environments. However, we were only able to examine a single dataset with just five different chunk sizes. Evaluating on more datasets and more chunk sizes would not only allow us to be more confident in our approach, but also would help investigate the presence of a scaling law for token-efficiency.

Additionally, we do not use long-context benchmarks such as RULER \citep{Hsieh2024RULERWT} and InfiniteBench \citep{Zhang2024BenchEL} as their tasks are mostly synthetic. Instead, we choose to evaluate with tasks that are grounded in real problems and of direct relevance to the community. Furthermore, the factors that these benchmarks aim to evaluate are retrieval, tracing, and aggregation, which we already evaluate using our tasks. Specifically, RepoQA requires retrieving exact code snippets from very large code repositories, LOFT-Spider requires tracing by resolving references across SQL tables, and BooookScore requires aggregation as it necessitates summarizing and synthesizing information across chunks. Thus, using additional benchmarks would introduce redundancy.

Another limitation of our work, and incremental memory approaches in general, is that they are less well-suited for precise multi-hop reasoning tasks. These tend to be synthetic tasks such as key-value tracing \citep{Liu2023LostIT}, where the context consists entirely of (key, value) pairs. Values may recursively point to other keys until a terminal value is found. If a value points to a key from a prior chunk, then for it to be traced successfully, this prior key-value pair must already be in the memory. To do so would require a complete and lossless memory as an incremental approach would not know in advance which precise key to keep in memory. This problem can be alleviated with multiple passes, reducing the usefulness of token efficiency, or by using a bidirectional or hierarchical approach. In realistic tasks that require tracing, such as LOFT-Spider where tables may refer to information from tables in prior chunks, we show that PRISM can still achieve reasonably strong performance, including outperforming other baselines.

Perhaps the most significant limitation is that, although we outperform other short context approaches, the ultimate goal is to achieve on-par performance with long-context models. Although we achieved this for the summarization task, there remains a gap to bridge for other tasks.

\subsubsection*{Acknowledgments}
Thanks to Michael Boratko and Zachary Fisher for comments and suggestions on this work. DJ was a Student Researcher at Google DeepMind during this project and is also supported by an AWS Studentship from the EPSRC Centre for Doctoral Training in Autonomous Intelligent Machines and Systems (AIMS) (EP/S024050/1).

\bibliography{custom}

\begin{thebibliography}{19}
\providecommand{\natexlab}[1]{#1}
\providecommand{\url}[1]{\texttt{#1}}
\expandafter\ifx\csname urlstyle\endcsname\relax
  \providecommand{\doi}[1]{doi: #1}\else
  \providecommand{\doi}{doi: \begingroup \urlstyle{rm}\Url}\fi

\bibitem[Chang et~al.(2024)Chang, Lo, Goyal, and Iyyer]{bookscore2024}
Y.~Chang, K.~Lo, T.~Goyal, and M.~Iyyer.
\newblock {BooookScore}: {A} systematic exploration of book-length summarization in the era of {LLM}s.
\newblock In \emph{The Twelfth International Conference on Learning Representations, {ICLR} 2024, Vienna, Austria, May 7-11, 2024}. OpenReview.net, 2024.
\newblock URL \url{https://openreview.net/forum?id=7Ttk3RzDeu}.

\bibitem[Fei et~al.(2024)Fei, Niu, Xie, Zhang, Bai, Deng, and Han]{retrievalreasoning2024}
W.~Fei, X.~Niu, G.~Xie, Y.~Zhang, B.~Bai, L.~Deng, and W.~Han.
\newblock Retrieval meets reasoning: Dynamic in-context editing for long-text understanding.
\newblock \emph{CoRR}, abs/2406.12331, 2024.
\newblock \doi{10.48550/ARXIV.2406.12331}.
\newblock URL \url{https://doi.org/10.48550/arXiv.2406.12331}.

\bibitem[He et~al.(2024)He, Qin, Prakriya, Sun, and Cong]{hmt2024}
Z.~He, Z.~Qin, N.~Prakriya, Y.~Sun, and J.~Cong.
\newblock {HMT:} hierarchical memory transformer for long context language processing.
\newblock \emph{CoRR}, abs/2405.06067, 2024.
\newblock \doi{10.48550/ARXIV.2405.06067}.
\newblock URL \url{https://doi.org/10.48550/arXiv.2405.06067}.

\bibitem[Hsieh et~al.(2024)Hsieh, Sun, Kriman, Acharya, Rekesh, Jia, and Ginsburg]{Hsieh2024RULERWT}
C.-P. Hsieh, S.~Sun, S.~Kriman, S.~Acharya, D.~Rekesh, F.~Jia, and B.~Ginsburg.
\newblock Ruler: What's the real context size of your long-context language models?
\newblock \emph{ArXiv}, abs/2404.06654, 2024.

\bibitem[Hwang et~al.(2024)Hwang, Zhou, Wendt, Gunel, Vo, Xie, and Tata]{cok2024}
E.~Hwang, Y.~Zhou, J.~B. Wendt, B.~Gunel, N.~Vo, J.~Xie, and S.~Tata.
\newblock Enhancing incremental summarization with structured representations.
\newblock In \emph{Findings of the Association for Computational Linguistics: EMNLP 2024}, pages 3830--3842, Miami, Florida, USA, Nov. 2024.
\newblock \doi{10.18653/v1/2024.findings-emnlp.220}.
\newblock URL \url{https://aclanthology.org/2024.findings-emnlp.220/}.

\bibitem[Hwang et~al.(2025)Hwang, Zhou, Gunel, Wendt, and Tata]{sumie2024}
E.~Hwang, Y.~Zhou, B.~Gunel, J.~B. Wendt, and S.~Tata.
\newblock {SUMIE}: A synthetic benchmark for incremental entity summarization.
\newblock In \emph{Proceedings of the 31st International Conference on Computational Linguistics}, pages 10839--10864, Abu Dhabi, UAE, Jan. 2025.
\newblock URL \url{https://aclanthology.org/2025.coling-main.721/}.

\bibitem[Ivgi et~al.(2023)Ivgi, Shaham, and Berant]{ivgi2023}
M.~Ivgi, U.~Shaham, and J.~Berant.
\newblock Efficient long-text understanding with short-text models.
\newblock \emph{Trans. Assoc. Comput. Linguistics}, 11:\penalty0 284--299, 2023.
\newblock \doi{10.1162/TACL\_A\_00547}.
\newblock URL \url{https://doi.org/10.1162/tacl\_a\_00547}.

\bibitem[Kwon et~al.(2023)Kwon, Li, Zhuang, Sheng, Zheng, Yu, Gonzalez, Zhang, and Stoica]{pagedattention2023}
W.~Kwon, Z.~Li, S.~Zhuang, Y.~Sheng, L.~Zheng, C.~H. Yu, J.~Gonzalez, H.~Zhang, and I.~Stoica.
\newblock Efficient memory management for large language model serving with pagedattention.
\newblock In \emph{Proceedings of the 29th Symposium on Operating Systems Principles, {SOSP} 2023, Koblenz, Germany, October 23-26, 2023}, pages 611--626. {ACM}, 2023.
\newblock \doi{10.1145/3600006.3613165}.
\newblock URL \url{https://doi.org/10.1145/3600006.3613165}.

\bibitem[Lee et~al.(2024)Lee, Chen, Dai, Dua, Sachan, Boratko, Luan, Arnold, Perot, Dalmia, Hu, Lin, Pasupat, Amini, Cole, Riedel, Naim, Chang, and Guu]{loft2024}
J.~Lee, A.~Chen, Z.~Dai, D.~Dua, D.~S. Sachan, M.~Boratko, Y.~Luan, S.~M.~R. Arnold, V.~Perot, S.~Dalmia, H.~Hu, X.~Lin, P.~Pasupat, A.~Amini, J.~R. Cole, S.~Riedel, I.~Naim, M.~Chang, and K.~Guu.
\newblock Can long-context language models subsume retrieval, rag, sql, and more?
\newblock \emph{CoRR}, abs/2406.13121, 2024.
\newblock \doi{10.48550/ARXIV.2406.13121}.
\newblock URL \url{https://doi.org/10.48550/arXiv.2406.13121}.

\bibitem[Liu et~al.(2024)Liu, Tian, Daita, Wei, Ding, Wang, Yang, and Zhang]{repoqa2024}
J.~Liu, J.~L. Tian, V.~Daita, Y.~Wei, Y.~Ding, Y.~K. Wang, J.~Yang, and L.~Zhang.
\newblock Repo{QA}: Evaluating long context code understanding.
\newblock \emph{CoRR}, abs/2406.06025, 2024.
\newblock \doi{10.48550/ARXIV.2406.06025}.
\newblock URL \url{https://doi.org/10.48550/arXiv.2406.06025}.

\bibitem[Liu et~al.(2023)Liu, Lin, Hewitt, Paranjape, Bevilacqua, Petroni, and Liang]{Liu2023LostIT}
N.~F. Liu, K.~Lin, J.~Hewitt, A.~Paranjape, M.~Bevilacqua, F.~Petroni, and P.~Liang.
\newblock Lost in the middle: How language models use long contexts.
\newblock \emph{Transactions of the Association for Computational Linguistics}, 12:\penalty0 157--173, 2023.

\bibitem[Munkhdalai et~al.(2024)Munkhdalai, Faruqui, and Gopal]{infiniattention2024}
T.~Munkhdalai, M.~Faruqui, and S.~Gopal.
\newblock Leave no context behind: Efficient infinite context transformers with infini-attention.
\newblock \emph{CoRR}, abs/2404.07143, 2024.
\newblock \doi{10.48550/ARXIV.2404.07143}.
\newblock URL \url{https://doi.org/10.48550/arXiv.2404.07143}.

\bibitem[Packer et~al.(2023)Packer, Fang, Patil, Lin, Wooders, and Gonzalez]{memgpt2024}
C.~Packer, V.~Fang, S.~G. Patil, K.~Lin, S.~Wooders, and J.~E. Gonzalez.
\newblock Mem{GPT}: Towards {LLM}s as operating systems.
\newblock \emph{CoRR}, abs/2310.08560, 2023.
\newblock \doi{10.48550/ARXIV.2310.08560}.
\newblock URL \url{https://doi.org/10.48550/arXiv.2310.08560}.

\bibitem[Pope et~al.(2023)Pope, Douglas, Chowdhery, Devlin, Bradbury, Heek, Xiao, Agrawal, and Dean]{scalinginference2023}
R.~Pope, S.~Douglas, A.~Chowdhery, J.~Devlin, J.~Bradbury, J.~Heek, K.~Xiao, S.~Agrawal, and J.~Dean.
\newblock Efficiently scaling transformer inference.
\newblock In \emph{Proceedings of the Sixth Conference on Machine Learning and Systems, MLSys 2023, Miami, FL, USA, June 4-8, 2023}. mlsys.org, 2023.

\bibitem[Qian et~al.(2024)Qian, Liu, Zhang, Mao, Zhou, Chen, and Dou]{Qian2024AreLA}
H.~Qian, Z.~Liu, P.~Zhang, K.~Mao, Y.~Zhou, X.~Chen, and Z.~Dou.
\newblock Are long-llms a necessity for long-context tasks?
\newblock \emph{ArXiv}, abs/2405.15318, 2024.

\bibitem[Reid et~al.(2024)Reid, Savinov, Teplyashin, Lepikhin, Lillicrap, Alayrac, Soricut, Lazaridou, Firat, Schrittwieser, Antonoglou, Anil, Borgeaud, Dai, Millican, Dyer, Glaese, Sottiaux, Lee, Viola, Reynolds, Xu, Molloy, Chen, Isard, Barham, Hennigan, McIlroy, Johnson, Schalkwyk, Collins, Rutherford, Moreira, Ayoub, Goel, Meyer, Thornton, Yang, Michalewski, Abbas, Schucher, Anand, Ives, Keeling, Lenc, Haykal, Shakeri, Shyam, Chowdhery, Ring, Spencer, Sezener, and others.]{gemini2024}
M.~Reid, N.~Savinov, D.~Teplyashin, D.~Lepikhin, T.~P. Lillicrap, J.~Alayrac, R.~Soricut, A.~Lazaridou, O.~Firat, J.~Schrittwieser, I.~Antonoglou, R.~Anil, S.~Borgeaud, A.~M. Dai, K.~Millican, E.~Dyer, M.~Glaese, T.~Sottiaux, B.~Lee, F.~Viola, M.~Reynolds, Y.~Xu, J.~Molloy, J.~Chen, M.~Isard, P.~Barham, T.~Hennigan, R.~McIlroy, M.~Johnson, J.~Schalkwyk, E.~Collins, E.~Rutherford, E.~Moreira, K.~Ayoub, M.~Goel, C.~Meyer, G.~Thornton, Z.~Yang, H.~Michalewski, Z.~Abbas, N.~Schucher, A.~Anand, R.~Ives, J.~Keeling, K.~Lenc, S.~Haykal, S.~Shakeri, P.~Shyam, A.~Chowdhery, R.~Ring, S.~Spencer, E.~Sezener, and others.
\newblock Gemini 1.5: Unlocking multimodal understanding across millions of tokens of context.
\newblock \emph{CoRR}, abs/2403.05530, 2024.
\newblock \doi{10.48550/ARXIV.2403.05530}.
\newblock URL \url{https://doi.org/10.48550/arXiv.2403.05530}.

\bibitem[Wang et~al.(2023)Wang, Dong, Cheng, Liu, Yan, Gao, and Wei]{wang2023}
W.~Wang, L.~Dong, H.~Cheng, X.~Liu, X.~Yan, J.~Gao, and F.~Wei.
\newblock Augmenting language models with long-term memory.
\newblock In \emph{Advances in Neural Information Processing Systems 36: Annual Conference on Neural Information Processing Systems 2023, NeurIPS 2023, New Orleans, LA, USA, December 10 - 16, 2023}, 2023.

\bibitem[Zhang et~al.(2024)Zhang, Chen, Hu, Xu, Chen, Hao, Han, Thai, Wang, Liu, and Sun]{Zhang2024BenchEL}
X.~Zhang, Y.~Chen, S.~Hu, Z.~Xu, J.~Chen, M.~K. Hao, X.~Han, Z.~L. Thai, S.~Wang, Z.~Liu, and M.~Sun.
\newblock Infinitebench: Extending long context evaluation beyond 100k tokens.
\newblock \emph{ArXiv}, abs/2402.13718, 2024.

\bibitem[Zheng et~al.(2023)Zheng, Yin, Xie, Huang, Sun, Yu, Cao, Kozyrakis, Stoica, Gonzalez, Barrett, and Sheng]{radix2023}
L.~Zheng, L.~Yin, Z.~Xie, J.~Huang, C.~Sun, C.~H. Yu, S.~Cao, C.~Kozyrakis, I.~Stoica, J.~E. Gonzalez, C.~W. Barrett, and Y.~Sheng.
\newblock Efficiently programming large language models using sglang.
\newblock \emph{CoRR}, abs/2312.07104, 2023.
\newblock \doi{10.48550/ARXIV.2312.07104}.
\newblock URL \url{https://doi.org/10.48550/arXiv.2312.07104}.

\end{thebibliography}

\clearpage
\onecolumn
\appendix

\section{Hand-crafted schemas}
\label{sec:handschema}
For BooookScore, the attributes map should be keyed by some identifier and the values should be a list of sentences that summarize the main plot of the book including events, background, themes, and characters. For RepoQA, each key is a function name and the value is a type that contains a natural language description of the function's purpose, input, output, and procedure. For LOFT-Spider, the schema maps table names to descriptions of the columns and relevant fields.

\begin{center}
\begin{minipage}{0.5\textwidth}
    \begin{lstlisting}[
        language=Python,
        basicstyle=\footnotesize\ttfamily, %
        breaklines=true,
        backgroundcolor=\color{white},
        caption=Schema for BooookScore.,
        label=lst:bookscore-schema
    ]
@dataclasses.dataclass
class BookSummary:
    """Keys may be whatever you want them to be. Values should summarize in sentences only the most important attributes of the book which should include absolutely essential details such as the main characters and their motivations, the main plot, the main events, background information, and the main theme."""
    attributes: dict[str, list[str]]
    \end{lstlisting}
\end{minipage}
\end{center}

\begin{center}
\begin{minipage}{0.8\textwidth}
    \begin{lstlisting}[
        language=Python,
        basicstyle=\footnotesize\ttfamily, %
        breaklines=true,
        backgroundcolor=\color{white},
        caption=Schema for RepoQA.,
        label=lst:repoqa-schema
    ]
@dataclasses.dataclass
class FunctionNaturalDescriptor:
    """candidate_functions is keyed by the exact name of the function and stores FunctionDescription objects. Each entry should represent a unique function, class method, property, getter, or setter (or anything defined with a def keyword) present in [TEXT] that potentially matches the description given in [QUESTION]. Never add the same function more than once and only add functions that appear similar to the description given in [QUESTION]. If two functions are very similar to each other, you should make sure to distinguish them in their FunctionDescription objects."""
  @dataclasses.dataclass
  class FunctionDescription:
    """purpose describes the purpose of the function i.e. what it does. input describes what the parameters of the function are. output describes what the function returns. procedure describes how the function is implemented (i.e. how it does what it does). Do not repeat the description given in [QUESTION]. You must describe the function based on what you see in [TEXT]."""
    purpose: str
    input: str
    output: str
    procedure: str
  candidate_functions: dict[str, FunctionDescription]
\end{lstlisting}
\end{minipage}
\end{center}

\begin{center}
\begin{minipage}{0.8\textwidth}
    \begin{lstlisting}[
        language=Python,
        basicstyle=\footnotesize\ttfamily, %
        breaklines=true,
        backgroundcolor=\color{white},
        caption=Schema for LOFT-Spider.,
        label=lst:loft-spider-schema
    ]
@dataclasses.dataclass
class RelevantTableInfo(pg.Object):
  """table_descriptions is a list of TableDescription objects. One for each table."""

  @dataclasses.dataclass
  class TableDescription(pg.Object):
    """Each element in columns_observed should provide the precise name and data type of a column in the table. In relevant_statistics you should calculate and provide statistics relevant to answering the query. relationships should provide the names of other tables that are related to this table and relevant to answering the query."""
    table_name: str
    table_description: str
    columns_observed: list[str]
    relevant_statistics: list[str]
    relationships: list[str]

  table_descriptions: list[TableDescription]
\end{lstlisting}
\end{minipage}
\end{center}

\section{LLM-generated schemas}
\label{sec:llmschema}

\begin{center}
\begin{minipage}{0.6\textwidth}
    \begin{lstlisting}[
        language=Python,
        basicstyle=\footnotesize\ttfamily, %
        breaklines=true,
        backgroundcolor=\color{white},
        caption=LLM-Generated Schema for BooookScore.,
        label=lst:llm-bookscore-schema
    ]
@dataclasses.dataclass
class NarrativeSummary:
  @dataclasses.dataclass
  class SummaryEvent:
    events: str
    characters: str
    places: str
    elements: str
  summary_events: dict[str, SummaryEvent]
\end{lstlisting}
\end{minipage}
\end{center}

\begin{center}
\begin{minipage}{0.5\textwidth}
    \begin{lstlisting}[
        language=Python,
        basicstyle=\footnotesize\ttfamily, %
        breaklines=true,
        backgroundcolor=\color{white},
        caption=LLM-Generated Schema for RepoQA.,
        label=lst:llm-repoqa-schema
    ]
@dataclasses.dataclass
class FunctionMatch:
  matches: dict[str, float]
\end{lstlisting}
\end{minipage}
\end{center}

\begin{center}
\begin{minipage}{0.5\textwidth}
    \begin{lstlisting}[
        language=Python,
        basicstyle=\footnotesize\ttfamily, %
        breaklines=true,
        backgroundcolor=\color{white},
        caption=LLM-Generated Schema for LOFT-Spider.,
        label=lst:llm-loft-spider-schema
    ]
@dataclasses.dataclass
class QueryPartialSolution(pg.Object):
  attributes: dict[str, list[str]]
\end{lstlisting}
\end{minipage}
\end{center}

\section{Using Typed Hierarchies in JSON}

\begin{figure}[h]
    \centering
    \includegraphics[width=1.0\linewidth]{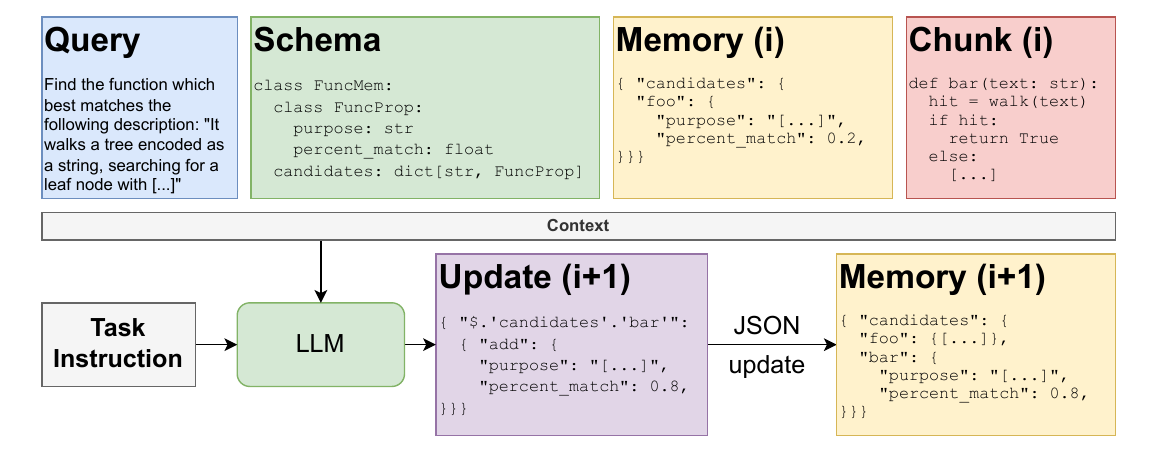}
    \caption{\textbf{Using a JSON-encoded memory.}
    Using the example of a code retrieval task, we fill the context of the LLM with the task instruction, query, a schema defining our memory structure, the existing memory, and a chunk of code context. When this context is used to prompt the LLM, it should propose a memory revision based on the chunk that it has seen. The revision is used to programmatically revise the underlying JSON memory structure, which is then used in the prompt for the next step.
    }
    \label{fig:json-overview}
\end{figure}

We present the use of Typed Hierarchies in JSON in figure \ref{fig:json-overview}. In this section, we refer to revisions as updates. The problem formulation is the same otherwise.

\label{sec:json}

\begin{figure}
    \centering
    \includegraphics[width=0.8\linewidth]{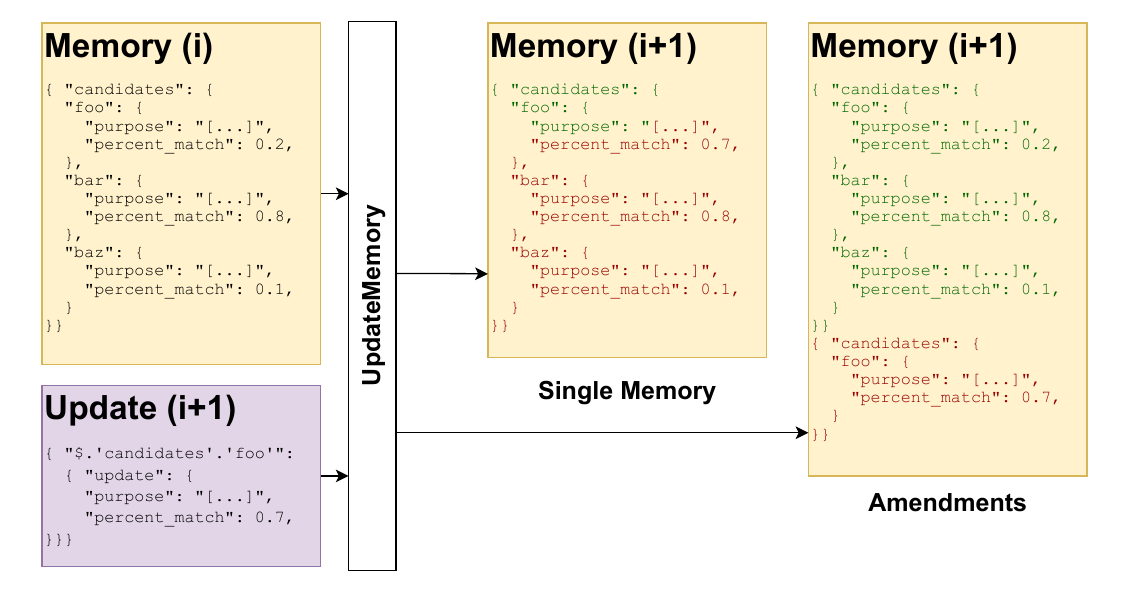}
    \caption{\textbf{Using amendments with a JSON memory.} An update to the candidate function ``foo'' is proposed, changing the existing ``percent\_match'' field from $0.2$ to $0.7$ and replacing the ``purpose'' field. The function \texttt{ReviseMemory} takes the existing memory and applies the proposed update to it. We show two possible resulting memory states. \textit{Single memory} shows the memory if the object at the specified path is updated with the new object directly. \textit{Amendments} shows the state if the change is simply amended to the end as a new JSON object. Text in green shows the longest matching prefix compared to the previous memory and text in red shows the information that must be (re-)encoded. Using amendments reduces the number of tokens that need to be encoded at the cost of increasing the size of the memory.
    }
    \label{fig:json-amendments}
\end{figure}

\section{Prompts}
\label{app:prompts}

In this section, we provide prompts for PRISM and LLM-assisted schema generation. Prompts for baselines may be found in \citet{bookscore2024}.

\begin{tcolorbox}[colback=blue!5!white, colframe=black!75!black, title=\center{\textbf{Example Prompt For PRISM}}, breakable, boxrule=0.5mm, colbacktitle=blue!15!white, coltitle=black]

\begin{lstlisting}[
    breaklines=true,
    frame=none,
    backgroundcolor=\color{blue!5!white}
]
{% raw %}I will provide a class definition [CLASS], which defines some fields that need to be generated, an instantiation of that class under [PARTIAL_SUMMARY] that is a response to the question in [QUESTION], and some text in a section called [TEXT]. Your task is to propose updates to [PARTIAL_SUMMARY] gathered from the information in [TEXT].

Here are the sections that you will complete, in the same order:

[OBJECTS FOR UPDATE]
In this section you will produce a set of dictionaries (one per line) in JSON format where the keys correspond go the JSONPaths whose objects should be updated and the values are a dictionary with the following fields:
* 'update': The object to overwrite the existing value at the JSONPath.

The 'update' object must adhere to the [CLASS] definition at the given JSONPath. If information is missing for some fields, then you can use the placeholder string '???' or `None` to represent that missing information. Updates must only ever increase the amount of information in [PARTIAL_SUMMARY], they can be made by either replacing a `None` object or the placeholder string '???' with relevant content from [TEXT] or by modifying an existing value using content from [TEXT]. To separate multiple values within a string, use '; '. For example, a value about the 'location' of a hotel may say '5 minute walk from the subway station; views of the Eiffel tower'.

[OBJECTS FOR ADD]
In this section you will produce a dictionary in JSON format where the keys correspond to the JSONPaths that do not exist in [PARTIAL_SUMMARY] yet that will be added, and the values are a dictionary with the following fields:
* 'add': The object to add at that JSONPath.

The 'add' object must adhere to the [CLASS] definition at the given JSONPath. If information is missing for some fields, then you can use the placeholder string '???' or `None` to represent that missing information.

Additional guidelines:

1. Field names in JSONPaths must be quoted. For example, "$.'places'.'flexibility of material'".
2. Proposed JSON objects must have sufficient context: the values of the [PARTIAL_SUMMARY] should have enough context so a reader can understand what it means.
3. Ignore irrelevant text: If the content in [TEXT] does not have any information that is relevant to [QUESTION] and [PARTIAL_SUMMARY] it is OK to make no update to that object.
4. No redundant fields: If information from [TEXT] can be incorporated by updating an existing field in [PARTIAL_SUMMARY], then do not introduce a new redundant field. For example, if there's already a field for 'activities' do not introduce a new field for 'other activities' or 'water activities', 'hiking'. Update the existing field for 'activities'.
5. A field should not contain redundant values. If one value encompasses most of the details in another value, merge them together. For instance, "beautiful views of the Eiffel tower" and "view of the Eiffel tower" should be merged into a single value like "beautiful views of the Eiffel tower".

Here is an example of an entity comparision:
[QUESTION]
ESR HaloLock wireless car charger vs. MagSafe Wireless Car Charger

[CLASS]
Comparison

```python
class Comparison:
  product_names: tuple[str, str]
  Facet = str
  values: dict[Facet, tuple[str, str]]
```

[PARTIAL_SUMMARY]
{
  "product_names": ("ESR HaloLock wireless car charger", "MagSafe Wireless Car Charger"),
  "values": {
    "size": ("4.2 inches", "???"),
    "flexibility": ("???", "very flexible"),
  }
}

[TEXT]
When using ESR HaloLock wireless car charger, the orientation of the iPhone is absolutely free, as they are free too inclination and height that you want to give to the smartphone, as long as you pay attention to the center of gravity: despite the aluminum structure it is very resistant and the joints are very solid, if you position the smartphone too far forward, the base of the stand would not be able to support the weight and would risk tilting forward.

[OBJECTS FOR UPDATE]
{"$.'values'.'flexibility'": {"update": ("allows iPhone to orientate freely", "very flexible")}}

[OBJECTS FOR ADD]
{"$.'values'.'material'": {"add": ("aluminum structure", "???")}}
{"$.'values'.'durability'": {"add": ("very resistant", "???")}}
{"$.'values'.'design'": {"add": ("can not support the weight of a phone if it is too far forward", "???")}}

===

Here is an example of an entity summarization:
[QUESTION]
Describe attributes and values of HOTEL0.

[CLASS]
class Summary(TypedDict):
  attributes: dict[str, list[str]]  # Keyed by attribute, with a list of sufficient details about the attribute.

[PARTIAL_SUMMARY]
{
  "attributes": {
    "Amenities": ["There are two pools", "pub opens till midnight"],
    "Food & Beverage": ["limited breakfast options"],
    "Room Quality": ["Spacious and comfortable rooms"],
  }
}

[TEXT]
HOTEL0 offers exceptional dining and the beds were very cozy, but there was a lot of street noise. HOTEL1 offers great Eiffel tower view from window.

[OBJECTS FOR UPDATE]
{"$.'attributes'.'Food & Beverage'": {"update": [ "limited breakfast options", "HOTEL0 offers exceptional dining" ]}}
{"$.'attributes'.'Room Quality'": {"update": [ "Spacious and comfortable rooms", "beds were very cozy" ]}}

[OBJECTS FOR ADD]
{"$.'attributes'.'Noise Level'": {"add": [ "Notable street noise at night" ]}}

===

Here is an example of query answering from SQL tables:
[QUESTION]
What is the total number of singers?

[CLASS]
class TableMemory(pg.Object):
  """table_descriptions is keyed by the name of the table and the value is a TableDescription object. One for each table."""

  class TableDescription(pg.Object):
    """table_description is a short summary of the table. thoughts is a list of relevant information from the table that will be helpful in answering the query. When referring to fields and columns, use their exact names."""
    table_description: str
    thoughts: list[str]

  table_descriptions: dict[str, TableDescription]

[PARTIAL_SUMMARY]
{
  "table_descriptions": {
    "Stadiums": {
      "table_description": "The table lists the stadiums including Wembley Stadium, Stark's Park, and others.",
      "thoughts": [
        "There are 8 stadiums in the table.",
        "It does not say which singers perform.",
      ],
    },
  },
}

[TEXT]
Table: Concert
concert_ID,concert_Name,Theme,Stadium_ID,Year
1,Auditions,Free choice,1,2014
2,Super bootcamp,Free choice 2,2,2014
3,Home Visits,Bleeding Love,2,2015
4,Week 1,Wide Awake,10,2014
5,Week 1,Happy Tonight,9,2015
6,Week 2,Party All Night,7,2015

Table: Singer
Singer_ID,Name,Country,Song_Name,Song_release_year,Age,Is_male
1,Joe Sharp,Netherlands,You,1992,52,F
2,Timbaland,United States,Dangerous,2008,32,T
3,Justin Brown,France,Hey Oh,2013,29,T
4,Rose White,France,Sun,2003,41,F
5,John Nizinik,France,Gentleman,2014,43,T
6,Tribal King,France,Love,2016,25,T

Table: Singer_in_Concert
concert_ID,Singer_ID
1,2
1,3
1,5
2,3
2,6
3,5
4,4
5,6
5,3
6,2

Table: Stadium
Stadium_ID,Location,Name,Capacity,Highest,Lowest,Average
1,Raith Rovers,Stark's Park,10104,4812,1294,2106
2,Ayr United,Somerset Park,11998,2363,1057,1477
3,East Fife,Bayview Stadium,2000,1980,533,864
4,Queen's Park,Hampden Park,52500,1763,466,730
5,Stirling Albion,Forthbank Stadium,3808,1125,404,642
6,Arbroath,Gayfield Park,4125,921,411,638
7,Alloa Athletic,Recreation Park,3100,1057,331,637
9,Peterhead,Balmoor,4000,837,400,615
10,Brechin City,Glebe Park,3960,780,315,552

[OBJECTS FOR UPDATE]
{}

[OBJECTS FOR ADD]
{"$.'table_descriptions'.'Singers'": {"add": {"table_description": "This table lists the singers.", "thoughts": [ "The Singer table has the singers Joe Sharp, Timbaland, Justin Brown, Rose White, John Nizinik, and Tribal King.", "There are 6 singers in the table." ]}}}

===

Here is an example of code retrieval:
[QUESTION]
Find the exact name of the function described by the following function descripition: 1. **Purpose**: The function generates a string used to format text with new lines and optionally a form feed character, typically used to control spacing in formatted output.
2. **Input**: The function takes three parameters: an integer representing the number of new lines, a boolean indicating whether a form feed character should be included, and an optional string representing the line break character (defaulting to a newline).
3. **Output**: It returns a string composed of the specified number of newline characters, and if requested, includes a form feed character followed by an additional newline.
4. **Procedure**: The function first checks if the form feed should be included. If true, it concatenates the specified number of newline characters minus one with a form feed character and another newline. If false, it simply returns a string of newline characters multiplied by the specified integer.

[CLASS]
class FunctionNaturalDescriptor(pg.Object):
  """candidate_functions is keyed by the exact name of the function and stores FunctionDescription objects. Each entry should represent a unique function, class method, property, getter, or setter (or anything defined with a def keyword) present in [TEXT] that potentially matches the description given in [QUESTION]. Never add the same function more than once and only add functions that appear similar to the description given in [QUESTION]. If two functions are very similar to each other, you should make sure to distinguish them in their FunctionDescription objects."""  # pylint: disable=line-too-long

  class FunctionDescription(pg.Object):
    """purpose describes the purpose of the function i.e. what it does. input describes what the parameters of the function are. output describes what the function returns. procedure describes how the function is implemented (i.e. how it does what it does). Do not repeat the description given in [QUESTION]. You must describe the function based on what you see in [TEXT]."""  # pylint: disable=line-too-long
    purpose: str
    input: str
    output: str
    procedure: str

  candidate_functions: dict[str, FunctionDescription]

[PARTIAL_SUMMARY]
{
  "candidate_functions": {
    "_merge_entities":{
      "purpose": "The function merges a list of entities into a single entity. Optionally, the merge can be done quickly, which may result in some information being lost. The function also formats and prints the merged entity.",
      "input": "The function takes two parameters: a list of entities to merge and a boolean indicating whether to do a quick merge (defaulting to False).",
      "output": "The function returns a single entity created from merging all entities in the input list.",
      "procedure": "The function first checks if the quick merge flag is set. If true, it uses a simple merge algorithm that simply concatenates all the entities in the list without any additional processing. If false, it uses a more complex merge algorithm that attempts to merge the entities in a way that preserves as much information as possible. The function then formats and prints the merged entity.",
    },
    "reduce_sum_list": {
      "purpose": "The function reduces a list of integers to a single integer by summing all the values in the list.",
      "input": "The function takes a single parameter: a list of integers.",
      "output": "The function returns a single integer representing the sum of all the values in the list.",
      "procedure": "The function iterates through the list of integers and adds each value to a running sum. It then returns the running sum.",
    }
  }
}

[TEXT]
File path: /src/test_file.py
Content:
        elif t in {token.NAME, token.NUMBER, token.STRING}:
            return NO

    elif p.type == syms.import_from:
        if t == token.DOT:
            if prev and prev.type == token.DOT:
                return NO

        elif t == token.NAME:
            if v == "import":
                return SPACE

            if prev and prev.type == token.DOT:
                return NO

    elif p.type == syms.sliceop:
        return NO

    elif p.type == syms.except_clause:
        if t == token.STAR:
            return NO

    return SPACE


def make_simple_prefix(nl_count: int, form_feed: bool, empty_line: str = "\n") -> str:
    """Generate a normalized prefix string."""
    if form_feed:
        return (empty_line * (nl_count - 1)) + "\f" + empty_line
    return empty_line * nl_count


def preceding_leaf(node: Optional[LN]) -> Optional[Leaf]:
    """Return the first leaf that precedes `node`, if any."""
    while node:
        res = node.prev_sibling
        if res:
            if isinstance(res, Leaf):
                return res

            try:
                return list(res.leaves())[-1]

            except IndexError:
                return None

        node = node.parent
    return None


def prev_siblings_are(node: Optional[LN], tokens: List[Optional[NodeType]]) -> bool:
    """Return if the `node` and its previous siblings match types against the provided
    list of tokens; the provided `node`has its type matched against the last element in
    the list.  `None` can be used as the first element to declare that the start of the
    list is anchored at the start of its parent's children."""

[OBJECTS FOR UPDATE]
{}

[OBJECTS FOR ADD]
{"$.'candidate_functions'.'make_simple_prefix'": {"add": {"purpose": "The function generates a prefix string by repeating a number of empty lines which can be formatted with a form feed character if supplied.", "input": "The function takes three parameters: an integer which denotes the number of new lines, a boolean determining if a form feed character should be used, and an optional argument: a string representing a character to be used for line breaks. This optional argument is defaulted to a newline character.", "output": "The function returns a string which is a prefix of the desired format.", "procedure": "The function first checks if the form feed character should be used. If it should, the function generates a string which is a concatenation of the specified number of newline characters minus one, a form feed character, and another newline character. If it should not, the function generates a string which is a concatenation of the specified number of newline characters. The function then returns the generated string." }}}
{"$.'candidate_functions'.'preceding_leaf'": {"add": {"purpose": "The function returns the first leaf that precedes a given node, if any.", "input": "The function takes a single parameter: a node to find the preceding leaf for.", "output": "The function returns a leaf if one exists, otherwise it returns None.", "procedure": "The function first checks if the node has a previous sibling. If it does, it checks if the previous sibling is a leaf. If it is, it returns the previous sibling. If it has a list of leaves, then it returns the last (or None if there is an IndexError). If it is not a leaf nor has a list of leaves, it sets the node to the parent and repeats the process in a while loop."  }}}
{"$.'candidate_functions'.'prev_siblings_are'": {"add": {"purpose": "The function checks if the node and its previous siblings match types against the provided list of tokens.", "input": "The function takes two parameters: a node to check and a list of tokens to match against.", "output": "The function returns a boolean indicating whether the node and its previous siblings match types against the provided list of tokens.", "procedure": "Unknown" }}}

===

{% endraw %}[QUESTION]
{{ question }}

[CLASS]
{{ structure }}

[PARTIAL_SUMMARY]
{{ partial_summary }}

[TEXT]
{{ text }}

{{ parse_response(store("raw_response", llm(temperature=0.8))) }}
\end{lstlisting}
\end{tcolorbox}

\begin{tcolorbox}[colback=blue!5!white, colframe=black!75!black, title=\center{\textbf{Example Prompt For Schema Generation (And Example Generations)}}, breakable, boxrule=0.5mm, colbacktitle=blue!15!white, coltitle=black]
\label{app:schemagen}

\begin{lstlisting}[
    breaklines=true,
    frame=none,
    backgroundcolor=\color{blue!5!white}
]
I have a system in which a user enters a query in a particular domain and the system must answer the query. In order to answer the query, the system views a large number of documents one at a time and only once. Thus, when the system sees a document it stores information related to answering the query by updating a JSON format memory. Once the system has viewed all documents, the system will read its JSON memory and use this information to decide the output for the query. Consequently, the JSON memory should store all information relevant to answering the query. Hence, the schema for the JSON memory must ensure that the memory contains the right information to assist the system in producing the final answer. It should make sure not to keep too little information nor too much-but generally should prefer to store more information if unsure. The schema for the memory is defined by a python dataclass and is specific to the domain of the query. The schema may include a python comment describing how it should be used. Your task is to construct a schema given a query domain and a query example.

The schema will always store some information from every document, so it should support data structures that can be appended or updated such as lists or dicts. Information that should be structured together should be kept together with subclasses.

===

[Query domain]
Comparing two entities found in various documents based on their respective shared attributes.

[Example query]
ESR HaloLock wireless car charger vs. MagSafe Wireless Car Charger

[Schema]
```python
class Comparison:
  """attributes should be keyed by attributes that both entities share e.g. connectivity and the values should be AttributeValues instances."""
  class AttributeValues:
    """entity_one and entity_two are lists of descriptions relating to the two respective entities, found in documents, that correspond to the specific attribute the instance is keyed under."""
    entity_one: list[str]
    entity_two: list[str]


  attributes: dict[str, AttributeValues]
```


===

[Query domain]
Summarizing details about entities (such as people, things, and institutions) found in online documents.

[Example query]
Describe attributes and values of HOTEL0.

[Schema]
```python
class Summary(TypedDict):
  """Keyed by attribute, with a list of sufficient details about the attribute."""
  attributes: dict[str, list[str]]
```

===

[Query domain]
Finding the top individuals in terms of a particular metric defined by the query. Documents are the content of websites on the internet.

[Example query]
Who are the top 10 highest earning CEOs in the bay area?

[Schema]
```python
class TopKList:
  """top_persons is a list of PersonMetric instances giving the person name and the value of the metric asked by the query."""
  class PersonMetric:
    """person is the name of the individual and metric is the value of the metric required by the query."""
    person: str
    metric: float


  top_persons: list[PersonMetric]
```

===

[Query domain]
Retrieving the exact name of a function given a query that describes the purpose, input, output, and procedure of the function. Documents are files of code. Here, the memory should provide some way of knowing to what extent a function matches the description given in a query.

[Example query]
Find the exact name of the function described by the following function description: 1. **Purpose**: The function generates a string used to format text with new lines and optionally a form feed character, typically used to control spacing in formatted output.
2. **Input**: The function takes three parameters: an integer representing the number of new lines, a boolean indicating whether a form feed character should be included, and an optional string representing the line break character (defaulting to a newline).
3. **Output**: It returns a string composed of the specified number of newline characters, and if requested, includes a form feed character followed by an additional newline.
4. **Procedure**: The function first checks if the form feed should be included. If true, it concatenates the specified number of newline characters minus one with a form feed character and another newline. If false, it simply returns a string of newline characters multiplied by the specified integer.


[Schema]

Generated Schema (for RepoQA):
class FunctionMatch:
  """Stores information about functions found in code."""
  class FunctionInfo:
    """name is the exact name of the function and matches is a list of strings describing which parts of the function description in the query were matched to the function."""
    name: str
    matches: list[str]


  functions: list[FunctionInfo]

===

[Query domain]
You are summarizing very long narrative books. Each document is a segment of the book. Here, the story may feature non-linear narratives, flashbacks, switches between alternate worlds or viewpoints, etc. Therefore, the memory needs to represent a consistent and chronological narrative. Critical information may relate to key events, backgrounds, settings, characters, their objectives, and motivations.

[Example query]
Summarize this book excerpt. Briefly introduce characters, places, and other major elements if they are being mentioned for the first time.

[Schema]

Generated Schema (for BooookScore):

class BookSummary:
  """Summarizes a book with potentially non-linear narratives by storing information chronologically."""
  class Event:
    """Represents a single event in the story. Events are stored in chronological order."""
    description: str
    """Description of the event."""
    time: str
    """Explicit time information provided in the text for this event, if any."""
    location: str
    """Location of the event, if specified."""
    characters: list[str]
    """Characters involved in the event."""

  class Character:
    """Represents a character in the story."""
    name: str
    """Name of the character."""
    description: str
    """Description or background information about the character."""
    motivations: list[str]
    """Known or speculated motivations of the character."""

  class Location:
    """Represents a location in the story."""
    name: str
    """Name of the location."""
    description: str
    """Description of the location."""

  events: list[Event]
  """List of events in the story, ordered chronologically."""
  characters: dict[str, Character]
  """Dictionary of characters encountered in the story, keyed by character name."""
  locations: dict[str, Location]
  """Dictionary of locations encountered in the story, keyed by location name."""

===

[Query domain]
Given a bunch of SQL tables formatted in text, answer queries that may require reasoning over multiple tables to find the answer.

[Example query]
What is the total number of singers?

[Schema]

LOFT-Spider generated schema:

class SQLQueryInformation:
    """This schema stores information relevant to a SQL query.
    It focuses on the entities and attributes mentioned in the query,
    rather than storing entire tables.
    """

    class EntityInformation:
        """Represents information about a specific entity mentioned in the query.
        For instance, if the query asks about 'singers', this would store
        information related to singers.
        """
        name: str  # Entity name (e.g., "singers")
        relevant_columns: list[str]  # Columns relevant to the query for this entity
        relevant_rows: list[dict[str, str]] # Rows containing information related to the query, as dictionaries

    entities: list[EntityInformation]

    # Additional fields for aggregate queries (COUNT, SUM, AVG, etc.):
    aggregate_results: dict[str, float] # e.g., {"count": 123}

\end{lstlisting}

\end{tcolorbox}

\section{Additional Results}
\label{app:more}

We present additional results, extending Table \ref{tab:efficiency} for the LOFT-Spider dataset in Table \ref{tab:moreeff}.

\begin{table}
    \centering
    \begin{tabular}{ll|ccccc}
    \toprule
    \multicolumn{2}{l|}{\textbf{Method}} & Cache hit & Enc. & Net Enc. & Dec. & Cost \\
    & & (\%) & ($\times10^3$) & ($\times10^3$) & ($\times10^3$) & Index \\
    \midrule
    & & \multicolumn{5}{c}{LOFT-Spider} \\
    \midrule
    \multicolumn{2}{l|}{Long context} & $0$ & $30$ & $30$ & $0.4$ & \cellcolor{gray!25} \\
    \midrule
    \multicolumn{2}{l|}{Incremental} & $0$ & $31$ & $\mathbf{31}$ & $1.5$ & $0.035$ \\
    \textbf{Ours} & In-place & $41$ & $56$ & $33$ & $0.7$ & $0.035$ \\
    & \quad\quad w/out updates & $39$ & $54$ & $33$ & $\mathbf{0.4}$ & $\mathbf{0.030}$ \\
    \textbf{Ours} & Amendments & $\mathbf{39}$ & $54$ & $33$ & $0.6$ & $0.035$ \\
    & \quad\quad w/out updates & $40$ & $54$ & $33$ & $\mathbf{0.4}$ & $0.034$ \\
    \bottomrule
    \end{tabular}
    \caption{\textbf{Cache-efficiency results for LOFT-Spider.}}
    \label{tab:moreeff}
\end{table}

\end{document}